\documentclass[10pt,twocolumn,letterpaper]{article}

\usepackage{cvpr}              
\usepackage{enumitem}
\usepackage{graphicx}
\usepackage{svg}
\usepackage{makecell}
\usepackage{multirow}

\definecolor{cvprblue}{rgb}{0.21,0.49,0.74}
\usepackage[pagebackref,breaklinks,colorlinks,citecolor=cvprblue]{hyperref}

\newcommand{\shortnamenospace}{GHNeRF}
\newcommand{\shortname}{GHNeRF }
\newcommand{\longname}{GHNeRF: Learning Generalizable Human Features\\with Efficient Neural Radiance Fields}

\def\paperID{8} 
\def\confName{CVPR}
\def\confYear{2024}

\title{GHNeRF: Learning Generalizable Human Features\\with Efficient Neural Radiance Fields}

\author{Arnab Dey\textsuperscript{1} \\
{\tt\small adey@i3s.unice.fr}
  \and
  Di Yang\textsuperscript{2}\\
{\tt\small di.yang@inria.fr}
  \and
  Rohith Agaram\textsuperscript{3}\\
{\tt\small rohithagaram@gmail.com}  
  \and 
  Antitza Dantcheva \textsuperscript{2}\\
  {\tt\small antitza.dantcheva@inria.fr}
  \and
  Andrew I. Comport\textsuperscript{1} \\
  {\tt\small andrew.comport@cnrs.fr}
  \and
  Srinath Sridhar\textsuperscript{4} \\ 
  {\tt\small srinath\_sridhar@brown.edu }
  \and
  Jean Martinet\textsuperscript{1}\\
  {\tt\small Jean.MARTINET@univ-cotedazur.fr}
\and 
  \\ I3S-CNRS/Universit\'e C\^ote d'Azur\textsuperscript{1}  $\>$ INRIA/Universit\'e C\^ote d'Azur\textsuperscript{2} 
  \\ IIIT Hyderabad\textsuperscript{3} $\>$ Brown University\textsuperscript{4}
}

\begin{document}
\maketitle

\begin{abstract}
Recent advances in Neural Radiance Fields (NeRF) have demonstrated promising results in 3D scene representations, including 3D human representations. However, these representations often lack crucial information on the underlying human pose and structure, which is crucial for AR/VR applications and games. In this paper, we introduce a novel approach, termed {\em \shortnamenospace}, designed to address these limitations by learning 2D/3D joint locations of human subjects with NeRF representation. \shortname uses a pre-trained 2D encoder streamlined to extract essential human features from 2D images, which are then incorporated into the NeRF framework in order to encode human biomechanic features. This allows our network to simultaneously learn biomechanic features, such as joint locations, along with human geometry and texture. To assess the effectiveness of our method, we conduct a comprehensive comparison with state-of-the-art human NeRF techniques and joint estimation algorithms. Our results show that \shortname  can achieve state-of-the-art results in near real-time. The project website \href{https://www.arnabdey.co/ghnerf.github.io/}{https://www.arnabdey.co/ghnerf.github.io/}.
\end{abstract}    
\begin{figure}[ht]
\begin{center}
   \includegraphics[width=0.9\linewidth]{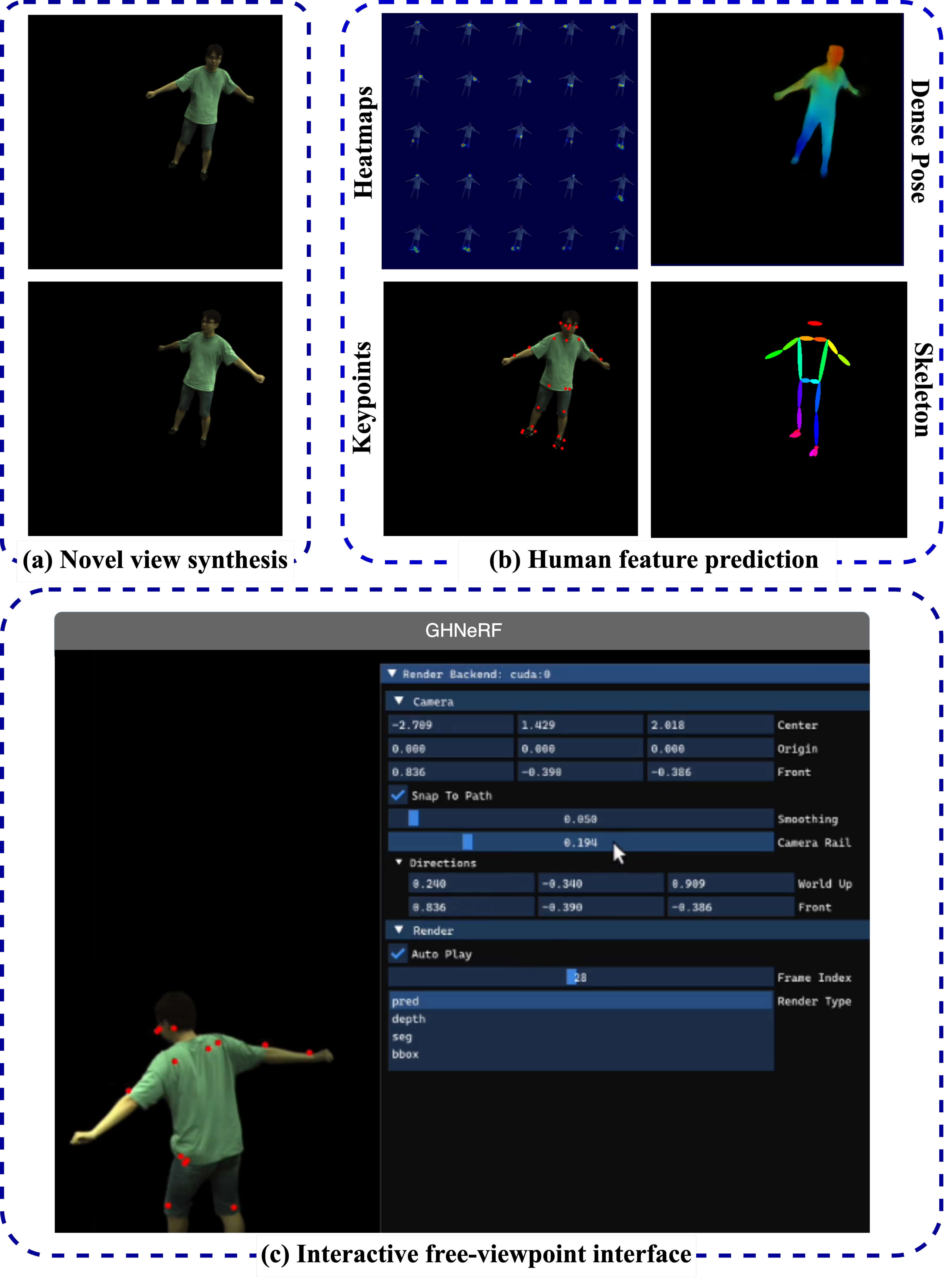}
\end{center}
\vspace{-15pt}
   \caption{In this work we propose \textbf{\shortnamenospace}, it can simultaneously learns both neural radiance fields and human features from sparse images. (a) shows high quality novel-view renderings. (b) shows generalizable human feature(keypoints, dense pose, etc.) estimated by \shortnamenospace. (c) present interactive tool to render free-viewpoint videos of novel-view and human features.}
   \vspace{-15pt}
\label{fig:banner}
\end{figure}
\section{Introduction}

\label{sec:intro} Developing a realistic virtual human model is pivotal for achieving natural experiences in Augmented Reality (AR) / Virtual Reality (VR) applications and interactive games. Moreover, creating custom photo-realistic virtual character from a sparse set of 2D images is one of the core challenges for AR/VR applications. Traditionally, this process has involved the use of elaborate multiview capture systems, which incorporate extensive camera arrays and body markers \cite{Dyna:SIGGRAPH:2015, h36m_pami}, to create human models with the underlying skeleton structure. These conventional methods predominantly utilize mesh representations, which are inherently constrained in terms of resolution and quality.  The underlying structures in these models are typically represented with parametric Skinned Multi-Person Linear (SMPL) models \cite{SMPL:2015} derived from body marker positioning. 
\par Recent advancements in Neural Radiance Fields (NeRF) have demonstrated remarkable potential in generating photorealistic virtual human avatars from mere 2D images \cite{hu2023sherf, su2021nerf, lin2022efficient}. However, existing NeRF-based approaches fall short in providing critical structural biomechanical attributes, crucial for various applications such as AR/VR, 3D animation, human performance analysis, and the medical field.
To bridge this gap, we introduce a Generalizable Human feature NeRF (\shortnamenospace), an end-to-end framework for learning generalizable human NeRF with biomechanic features. Human biomechanics refers to the study of human movement focusing musculoskeletal system, comprising bones, muscles, ligaments, and joints~\cite{lu2012biomechanics}. Within the scope of this paper, the term 'biomechanical features' specifically refers to the bones, muscles, and joints integral to this system.
Deviating from previous methods such as PixelNeRF~\cite{yu2021pixelnerf}, which used a 2D encoder to learn generalizable NeRF for view synthesis, our approach utilizes 2D deep feature extractors to simultaneously learn human features with generalizable NeRF models. Here, we demonstrate that it is possible to learn 3D human features from 2D images using the NeRF architecture. The \shortname predicts human features, such as heatmaps, facilitating 2D/3D joint estimation for novel views, which are applied to various downstream applications. We highlight that while we focussed on the \textit{ joint prediction}, the architecture can be used to learn other biomechanic properties, such as body part segmentation.
\par Our methodology adopts a 2D encoder similar to previous methods~\cite{yu2021pixelnerf, ye2023featurenerf} aimed at generating pixel-aligned human features from images. For this purpose, we compare two types of encoder inspired by previous state-of-the-art pose estimation algorithms. \shortname determines heatmaps corresponding to each joint, along with the color and volume density for each 3D query point. The input for the MLP are pixel-aligned features from encoder, as well as view direction. The heatmaps are generated using volume rendering similar to rendering color in NeRF. We use an efficient and generalizable NeRF architecture as a backbone similar to the one presented by Lin \textit{et al.} \cite{lin2022efficient} that allows for near real-time inference. 
\par To evaluate \shortnamenospace, we present the result of keypoint estimation tasks using two popular datasets. To our knowledge, our method is the first to provide human biomechanic features from NeRF. 
Our contributions are summarized as follows.
\begin{itemize}[noitemsep]
    \item We introduce \shortnamenospace, a novel generalizable NeRF architecture capable of accurately estimating 2D/3D human keypoints.
    \item \shortname demonstrates the ability not only to predict human keypoints but also to estimate complex human features, such as dense poses. This capability can also be achieved through the distillation of SoTA pose estimation algorithms.
    \item We provide a generalizable approach for predicting human feature, photometric, and geometric representations from 2D sparse images, applicable in interactive, real-time applications.
    \item We conduct extensive experimental analyses across various types of human images using two distinct datasets to validate the applicability and versatility of \shortname.  
\end{itemize}





\section{Releted works}
The proposed \shortname uses sparse multiview images of different humans to learn a generalizable NeRF representation that can also produce a consistent 3D human feature without any prior supervision during inference time. In the following, work related to this research will be discussed.

\subsection{NeRF for 3D representation}
In recent years, the NeRF-based method has gained significant popularity for the visual quality of 3D scene representations. NeRF~\cite{mildenhall2021nerf} represents 3D scenes using MLP by mapping 3D coordinates and 2D view directions to density and color. The original paper \cite{mildenhall2021nerf} and the following research work \cite{barron2021mip,barron2022mip,martin2021nerf, pumarola2021d, park2021nerfies} showed the effectiveness of the neural field compared to other classical methods for representing 3D and 4D scenes. The works \cite{yu2021plenoctrees, muller2022instant, lin2022efficient, Dey2022B02} address the long training and inference time of the NeRF by using faster sampling techniques, voxel representation, and hash encoding. Another limitation of NeRF-based methods is that they are scene specific, PixelNeRF~\cite{yu2021pixelnerf} showed that NeRF models can be generalized by conditioned NeRF on input image. More recently, FeatureNeRF~\cite{ye2023featurenerf} learned deep features using pre-trained vision foundation models for downstream applications such as semantic segmentation and key point transfer. Several methods~\cite{zhi2021place, kundu2022panoptic, wang2022dm} extended the NeRF's ability by learning scene properties, for example, semantic segmentation of the scene. However, most of the previous work focuses on scene features, such as segmentation. Our work differs from them by learning human biomechanic features with NeRF.

\subsection{NeRF for human representation}
In recent research, Hu et al. \cite{hu2023sherf} generated genralizable and animatable human NeRF models from a single input image. Although they achieved great results, their method relies on the SMPL parameters as input along with image, which is difficult to obtain in a real-world scenario. Similarly, GM-NeRF~\cite{chen2023gm} used the SMPL model to learn a generalizable human NeRF model. Several works~\cite{xu2021h, yu2023monohuman, jiang2023instantavatar} generated NeRF models of human in canonical T-pose (example SMPL~\cite{loper2023smpl} T-pose) then map it to a posed space. Similarly, \cite{peng2023implicit, jiang2022neuman, weng2022humannerf, su2021nerf} uses pre-existing skeleton data or pose estimator or information from the SMPL model~\cite{loper2023smpl} to reconstruct novel views or novel poses. As an example, A-NeRF~\cite{su2021nerf} employs off-the-shelf pose estimators to initialize their model, while our generalizable method does not require any pose initialization. In this paper, we predict human biomechanic features, such as joint information, directly from 2D images without any supervision.
 \begin{figure*}[t]
\begin{center}
   \includegraphics[width=1\linewidth]{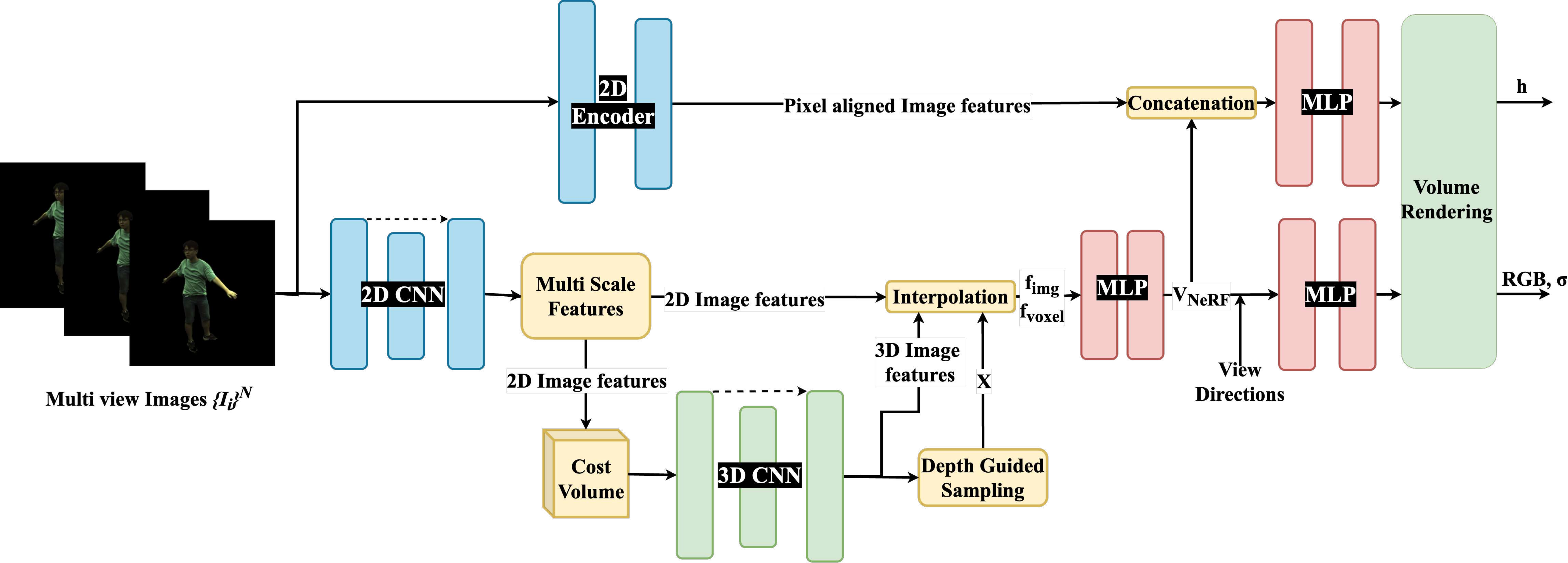}
\end{center}
   \caption{\textbf{Overview of the \shortname pipeline:} Given an input image $I$, human features $f_\textbf{h}$ and multi-resolution image features $f_{img}$ can be extracted using a 2D image encoder and a 2D CNN respectively. Subsequently, $f_{img}$ is used to form a cost volume for depth prediction. The predicted depth is used for depth-guided sampling to reduce the number of samples along the ray. For each 3D sample point $x$ along the ray, we combine image and voxel features to input an MLP $g_{NeRF}$, generating the intermediate NeRF feature $V_{NeRF}$. Finally, the intermediate NeRF feature $V_{NeRF}$ and the human feature $f_\textbf{h}$ are concatenated and fed into a smaller MLP $g_h$ to produce heatmaps. Furthermore, $V_{NeRF}$ and the view direction $\textbf{d}$ are combined in another MLP $g_c$ to derive color $c$. The final pixel color and heatmaps are generated using volume rendering technique.}
\label{fig:method}
\end{figure*}
\subsection{Human pose estimation}
Human pose estimation has been a long-standing problem in computer vision for decades. Most state-of-the-art approaches for 2D human pose estimation employ 2D CNN architectures for a single image in a strongly supervised setting~\cite{ning2017knowledge, OpenPose, Wang2018MagnifyNetFM, fang2017rmpe,He_2017_ICCV,cheng2020bottom, Kreiss_2019_CVPR}. For 3D pose estimation, \cite{RogezWS18, Moon_2019_ICCV_3DMPPE} focus on end-to-end reconstruction by directly estimating 3D poses from RGB images without intermediate supervision. \cite{zhaoCVPR19semantic} applies GCNs for regression tasks, especially 2D to 3D human pose regression. \cite{pavllo:videopose3d:2019} demonstrates that 3D poses in video can be effectively estimated with a fully convolutional model based on dilated TCNs over 2D keypoint sequences. Among these methods, \cite{RogezWS18, ning2017knowledge, Moon_2019_ICCV_3DMPPE,Wang2018MagnifyNetFM}
have first incorporated a person detector, followed by the estimation of the joints and then the computation of the pose for each person -- 
however the detection speed is proportional to the number of people in the image. 
Bottom-up methods such as \cite{OpenPose, cheng2020bottom, Kreiss_2019_CVPR} detect joints via heatmaps and associate body parts, but struggle with occluded or truncated body parts. Our approach integrates an encoder with NeRF to directly estimate heatmaps from 3D NeRF features, enhancing accuracy in predicting non-visible regions in 2D.

\section{Method}
We present \shortnamenospace, a unified framework for learning generalizable human features with the efficient NeRF architecture.
First, we present an introduction to NeRF and its generalizable variants. Then in Section~\ref{humanfeature},
we outline the feature extraction process and explain how to learn human features with NeRF in Section~\ref{learningNeRF}. Finally, we provide details of keypoint extraction in Section~\ref{estimation} .

\subsection{Preliminaries}
Neural Radiance Fields (NeRF) learn 3D scene representations using a multilayer perceptron (MLP). The input to the MLP consists of 3D coordinates $x = (x, y, z)$ and the view direction $\textbf{d} = (\theta, \phi)$. The outputs are color, $c = (r, g, b)$ and density$(\sigma)$. It can be represented as: \(F (x, \textbf{ d}) \rightarrow (c, \sigma)\) then volume rendering is used to generate the final pixel colors from the output. To predict images, first, 3D points are sampled along the rays $r(t)=o+t\textbf{d}$ passing through each pixel, with $o$ the camera center and $\textbf{d}$ the direction of the ray. The color and density of the samples are predicted using an MLP as discussed before. The final color of the pixel $C$ of a camera ray $r(t)=o+t\textbf{d}$ can be calculated as:
\begin{equation}
    C(r) = \int_{t_n}^{t_f} T(t)\tau(r(t))c(r(t),\textbf{d})dt,
\end{equation} where \(
    T(t)=exp(-\int_{t_n}^{t} \tau(r(s))ds).\)
The function $T(t)$ denotes the accumulated transmittance along the ray from $t_n$ to $t$, $t_n$ and $t_f$ is near and far bound of the ray. In practice, the color $\widehat{C}(r)$ is estimated by obtaining discrete samples along the ray, and the integral is approximated using numerical quadrature techniques.


In case of generalizable NeRF, the NeRF models are conditioned on the input image $I$:
\begin{equation}
\begin{gathered}
    \sigma(x, I) = g_\sigma(x, f(I)_{\pi(x)}) \\
    c(x,\textbf{d},I) = g_c(x,\textbf{d},f(I)_{\pi(x)}),
\end{gathered}
\end{equation}
where $g_\sigma$ and $g_c$ are two MLPs that predict density and color, $f$ is an image encoder and $\pi$ is a projection function that projects $x$ into the image plane using the known pose and intrinsic. The image passes through an encoder to generate features, then for each query point $x$, the corresponding pixel-aligned features~\cite{yu2021pixelnerf} $f(I)_{\pi(x)}$ are concatenated with the positional encoding of the point before inputting into the NeRF model. Similarly, ENeRF~\cite{lin2022efficient} extracts multiscale image features from a CNN- based encoder, and then the encodings are also used as input and to create a cost volume. Given the cost volume, a 3D CNN generates a depth probability volume, which is used to predict the depth probability of a pixel. ENeRF uses depth probabilities to sample points close to the surface, resulting in fewer samples and faster training and inference time. 

%
\subsection{Feature extraction}
\label{humanfeature}
We propose a new architecture to generalize human NeRF with the underlying biomechanic features. The original NeRF model predicts the color $c$ and the density $\sigma$ for each query point $x=(x,y,z)$, while the most generalizable NeRF models are conditioned on input images. We take inspiration from previous generalizable methods~\cite{yu2021pixelnerf, ye2023featurenerf, lin2022efficient}, and we use two different encoders: one to generate a human feature and the other for multiscale image features similar to \cite{lin2022efficient}. Each query point is projected on the input images, and then the pixel-aligned image features from each image are combined using a pooling operator~\cite{lin2022efficient} that is denoted as $f_{img} = \psi(f_1,...,f_N)$ where $f_N$ represents the feature of the $N^{th}$ image. Multiscale features are also used to generate voxel-aligned features similar to~\cite{lin2022efficient} denoted by $f_{voxel}$. Subsequently, we introduce a second encoder to encode human features. It has been demonstrated that human features extracted from Transformer-based encoder~\cite{dosovitskiy2020vit} pre-trained on ImageNet are more effective in generalizing human pose estimation~\cite{xu2022vitpose, zheng2021poseformer} compared to CNN-based features~\cite{RogezWS18, alphapose}. In this work, we compare both types of encoders and select the vision transformer encoder~\cite{dosovitskiy2020vit} to extract more effective features for human pose estimation. Specifically, we use a pre-trained vision transformer to extract a higher-dimension feature vector $\textbf{h}$ following~\cite{caron2021emerging}. For each query point $x$, we combine all pixel-aligned human features $f_{h} = \psi(\textbf{h}_1,...,\textbf{h}_N)$ from input images with a pooling operator. 
\subsection{Learning human features with NeRF}
\label{learningNeRF}
Gerneralizable NeRF models predict color $c$ and $\sigma$ for any query points, \shortname extends the generalizable NeRF models to predict additional features, in this case human joint locations. Although we have extracted features from images, we still need to incorporate them with NeRF, in order to output 3D consistent human features from NeRF. In this work, we learn intermediate NeRF features $V_{NeRF}(x,I)$ similar to \cite{ye2023featurenerf}. Then we use a number of small MLPs to predict other outputs from the intermediate NeRF feature:
\begin{equation}
\begin{gathered}
    V_{NeRF}(x, I) = g_{NeRF}(f_{img}, f_{voxel}) \\
    \sigma(x, I) = g_\sigma(V_{NeRF}(x, I)) \\
    c(x,\textbf{d},I) = g_c(V_{NeRF}(x, I), \textbf{d}) \\
    h(x, I) = g_h(V_{NeRF}(x, I), f_\textbf{h}).\\
\end{gathered}
\end{equation}
The color is predicted using a smaller MLP $g_c$ that takes the intermediate NeRF input feature $V_{NeRF}$ and the view direction as input. 
An additional branch predicts human joint locations as heatmaps $h$ from NeRF features. We take intermediate NeRF features before outputting color and density and concatenate with human feature $f_\textbf{h}$ and pass it through a smaller MLP $g_h$ that outputs heatmaps as feature vector $h\in\Re^J$ where $J$ is the number of joints. We can aggregate these feature vectors along the rays similar to color using volume rendering:
\begin{equation}\widehat{H}(r) = \sum_{i=1}^{N}T_i(1-exp(-\tau_i\delta_i))h_i,\end{equation} where \(h_i = g_h(V_{NeRF}(x, I), f_h)\) and $V_{NeRF}(x,I)$ denotes intermediate NeRF features.
The network is optimizing using a set of human images in a random pose and appearance with known camera parameters. The proposed method is optimized using photometric and feature loss. The photometric loss $l_{col}$ is calculated using the mean squared error between the predicted and the ground-truth color. We also add perceptual loss $l_{perc}$ to image patches similar to \cite{lin2022efficient}. Feature loss $l_{heat}$ is the mean square error between the predicted feature and the ground-truth feature in this case heatmaps. 
The final loss function can be represented as:
\[l = l_{col} + \lambda_p l_{perc} + \lambda_h l_{heat}\] where $\lambda_p$, $\lambda_h$ weighting coefficients.
During training, when ground truth features are not present, our method represents a student network, which can learn heatmaps through distillation of advanced heatmap-based pose estimation algorithms. The pose estimation algorithm~\cite{OpenPose} acts as a teacher network with the ability to predict heatmaps, thus guiding our student network in its heatmap prediction task.

\begin{figure*}[ht!]
    \centering
     \includegraphics[width=1\linewidth]{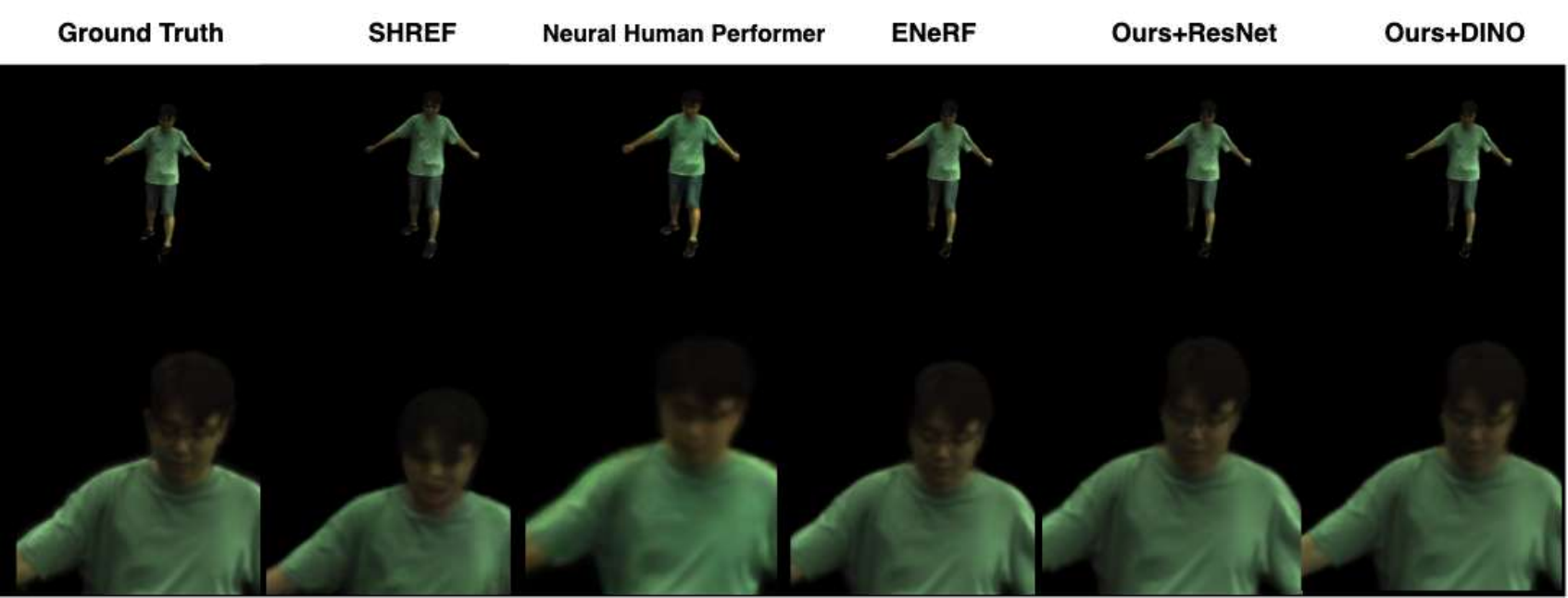}
    \caption{Qualitative comparison of generalization results on ZJU\_MoCap unseen test sequence. }
    \label{fig:method_compare}
\end{figure*}
\subsection{Keypoints extraction}
\label{estimation}
The 2D keypoint locations are estimated from the predicted heatmaps generated by NeRF. We calculate the 2D keypoints in a similar way to OpenPose~\cite{OpenPose}. A Gaussian filter is applied to the heatmaps, and then each channel is converted to a binary map by applying a threshold. Connected regions are created from binary maps, and the peak value within that region is calculated. The pixel with the peak value is then outputted as the 2D keypoint. To extract the 3D keypoints, we query sample points from a 3D volume around the subject and extract a volumetric heatmap. The 3D keypoints are calculated from the volumetric heatmap in a similar way as in the 2D keypoints.


\section{Experimental Results}
We conducted a thorough evaluation of the ability of our model to learn human features, particularly to estimate human joint locations. We carried out extensive experiments on two distinct datasets and compared our results with those of other leading human NeRF techniques.
\subsection{Experimental Setting}
\textbf{Datasets:} We trained our model to be applicable to various types of human image using two different datasets, namely ZJU\_MoCap \cite{peng2021neural} and RenderPeople \cite{hu2023sherf}. Both datasets are focused on humans and contain dynamic sequences of different individuals performing various activities. The ZJU\_MoCap dataset contains real images, while the RenderPeople dataset contains simulated images. ZJU\_MoCap includes 9 dynamic sequences (images, masks, camera parameters, and 2D/3D joint locations) of 9 different individuals performing 9 different actions. We randomly divided 6 sequences for training and 2 for testing and removed one sequence due to missing frame data. For RenderPeople, we randomly chose 440 sequences for training and 60 for testing.\\
\begin{figure*}[ht!]
    \centering
     \includegraphics[width=0.9\linewidth]{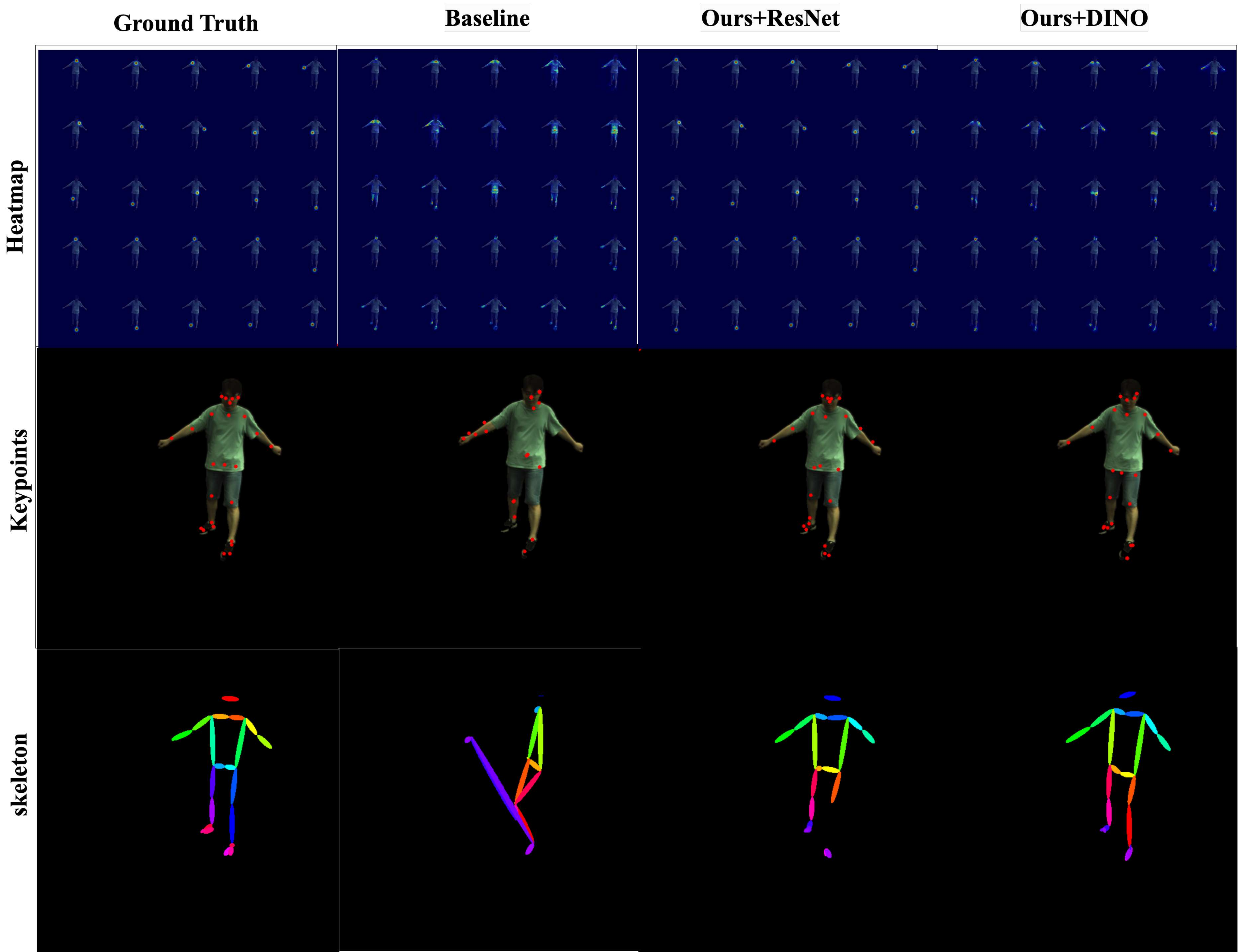}
    \caption{\small Qualitative result of keypoint estimation on ZJU\_MoCap dataset.}
    \label{fig:keypoint}
\end{figure*}
\textbf{Baseline:} We predominantly compare \shortname with other methods based on dynamic human NeRF. Although such methods are generalizable, none are capable of generating human features. We have extended ENeRF\cite{lin2022efficient} to output heatmaps by adding an additional output branch and reported its performance as a baseline for the joint estimation task.\\
\textbf{Implementation details:} We employ ENeRF as the base generalizable NeRF architecture due to its efficiency and generalizability, and proceeded to modify it to generate generalizable human features. We employed two distinct encoders, ResNet~\cite{he2016deep} and DINO~\cite{caron2021emerging}, in accordance with the most recent pose estimation techniques. 
We implemented our generalizable NeRF model using PyTorch. 
We trained the models with an RTX 3090 GPUs, using the Adam optimizer with an initial learning rate of $5e^{-4}$. We halved the learning rate every 50k iterations, and the model generally converged after about 200k iterations, taking about 18 hours. 
The weights of different losses are $\lambda_h=0.5$ and $\lambda_p=0.01$. For more information on the network architecture and other implementation details, see the Supplementary Material.\\
\textbf{Metrics:} 
We employed five different metrics to evaluate the predicted RGB image, heatmaps, and joint estimation quality. Peak Signal-to-Noise Ratio (PSNR in dB): To compare the quality of the RGB reconstruction, the higher is better; Structural Similarity Index (SSIM): To compare image quality in the reconstructed image, the higher is better; Learned Perceptual Image Patch Similarity (LPIPS)~\cite{zhang2018unreasonable}: the distance between the patches of the image, the lower means that the patches are more similar; Mean Squared Error (MSE): Mean squared distance between ground truth heatmap and predicted heatmap, lower the better; Percentage of correct keypoints (PCK): Measures whether the predicted keypoints and the true joint are within a certain distance threshold. We use PCK@0.2: Distance between the predicted and true joint $< 0.2 \times$ torso diameter. 

\subsection{Performance on novel view synthesis and joint estimation}
\begin{table*}[tbh!]
    \centering
    \begin{tabular}{c|ccccc}
    \hline
         Method & PSNR $\uparrow$ & SSIM $\uparrow$ & LPIPS $\downarrow$ & MSE $\downarrow$ & PCK $\uparrow$ \\
    \hline
         SHERF~\cite{hu2023sherf} & 26.37 & 0.918 & 0.1023 & - & - \\
         \makecell{Neural Human\\Performer~\cite{kwon2021neural}} & 25.76 & 0.906 & 0.148 & - & - \\
         ENeRF~\cite{lin2022efficient} & 31.48 & 0.965 & \textbf{0.0494} & - & - \\
         ENeRF+Heatmap & 31.48 & 0.965 & 0.050 & 0.0005 & 0.438\\
         Ours+ResNet & 31.20 & 0.963 & 0.054 & 0.0004 & 0.573\\
         Ours+DINO & \textbf{31.61} & \textbf{0.966} & 0.050 & \textbf{0.0003} & \textbf{0.691}\\
    \hline
    \end{tabular}
    \caption{Quantitative comparison of generalization (unseen test set) on the ZJU\_MoCap dataset, evaluating all methods at $512\times512$ resolution. For these experiments, we adhered to the default configurations of SHERF, Neural Human Performer, and ENeRF.}
    \label{tab:compare_method}
\end{table*}
\begin{table}[tbh!]
    \centering
    \scalebox{0.8}{
    \begin{tabular}{c|ccccc}
    \hline
        Dataset & PSNR$\uparrow$ & SSIM$\uparrow$ & LPIPS$\downarrow$ & MSE$\downarrow$ & PCK$\uparrow$\\
    \hline
        ZJU\_MoCap+Res& 31.20 & 0.963 & 0.054 & 0.0004 & 0.573\\
        ZJU\_MoCap+DINO& 31.61 & 0.966 & 0.050 & 0.0003 & 0.691\\
        RenderPeople+Res & 34.44 & 0.992 & 0.0131 & 0.0012 & 0.521\\
        RenderPeople+DINO & 34.75 &  0.992 & 0.0131 & 0.0005 & 0.502\\
    \hline
    \end{tabular}
    }
    \caption{Quantitative results of the proposed method in different datasets. The results represent generalizable performance on unseen scenes from the test set. Both datasets are evaluated on images with resolution $512 \times 512$.}
    \label{tab:compare_dataset}
\end{table}
We compared our method with recent generalizable NeRF-based methods on dynamic scenes, Table:~\ref{tab:compare_method} lists the quantitative result on ZJU\_MoCap dataset, which shows our method achieves state-of-the-art performance, while additionally estimating human joints. To establish a baseline, we incorporated an additional heatmap breach into ENeRF. The experiments show that our method maintains the same level of performance in novel-view synthesis compared to state-of-the-art ENeRF~\cite{lin2022efficient} but performs significantly better in joint estimation compared to the baseline ENeRF. It also demonstrates that the human feature encoder offers essential information about human features to more accurately estimate heatmaps crucial for better joint estimation. 
Figure~\ref{fig:method_compare} illustrates the qualitative outcomes of various approaches in ZJU\_MoCap dataset. Our technique demonstrates highly competitive results in novel-view synthesis and notably outperforms SHREF~\cite{hu2023sherf} and the Neural Human Performer~\cite{kwon2021neural} in preserving intricate details..

\begin{figure}[ht!]
    \centering
     \includegraphics[width=1\linewidth]{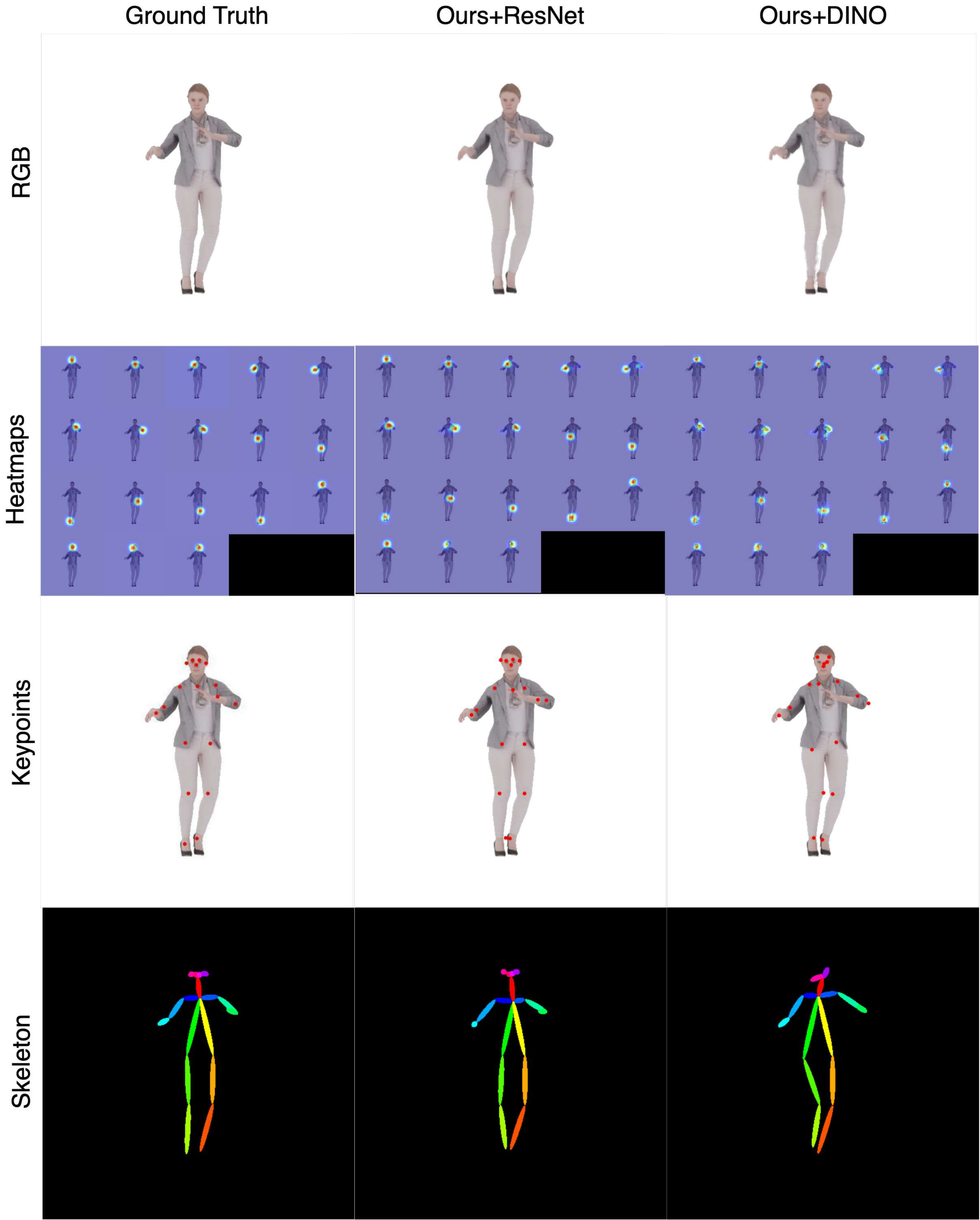}
    \caption{\small Qualitative result of keypoint estimation on RenderPeople dataset.}
    \label{fig:renderpeople}
\end{figure}
\begin{table}[tbh!]
    \centering
    \scalebox{1}{
    \begin{tabular}{c|cccc}
    \hline
         & \makecell{Alpha\\Pose~\cite{alphapose}} & \makecell{Open\\Pose~\cite{OpenPose}} & \makecell{\shortname\\+Res} & \makecell{\shortname\\+DINO}\\
    \hline
       PCK $\uparrow$  & 0.647 & 0.632 & 0.573 & \textbf{0.691} \\
       MSE $\downarrow$  &  0.0013 & 0.0015 & 0.0004& \textbf{0.0003}\\
    \hline
    \end{tabular}}
    \caption{Quantitative results of keypoint estimation compare to other pose estimation algorithms. We used same ZJU\_MoCap test set images of resolution $512 \times 512$ to evaluate all three methods.}
    \label{tab:compare_pose}
\end{table}
 In Figure~\ref{fig:keypoint}, we have presented qualitative results of human joint estimation task using the ZJU\_MoCap dataset. We generated 25 distinct heatmaps representing different keypoints, with each keypoint being highlighted by red markers. We evaluated our approach using two different types of dataset and reported quantitative results in Table~\ref{tab:compare_dataset}. In both datasets, the DINO features showed superior performance in predicting human features as heatmaps. We validate our approach using both real images and simulated images to demonstrate its robustness. Qualitative results of novel-view synthesis and joint estimation on RenderPeople dataset are presented in Figure:~\ref{fig:renderpeople}. To gauge the effectiveness of our proposed method for joint estimation, we compared it with other state-of-the-art pose estimation algorithms and presented the findings in Table~\ref{tab:compare_pose}. Our approach with both ResNet and DINO encoder outperform Alpha Pose and Open Pose, achieving superior PCK and MSE scores. More details, experiments, and results are provided in the Supplementary Material. 
\begin{table}[h!]
    \centering
    \scalebox{0.9}{
    \begin{tabular}{c|cccc}
    \hline
        Dataset & PSNR$\uparrow$ & SSIM$\uparrow$ & LPIPS$\downarrow$ & MSE $\downarrow$\\
    \hline
        ZJU\_MoCap+Res&  37.22 & 0.9885 & 0.0190 & 0.0039\\
        ZJU\_MoCap+DINO& 36.51 & 0.9877 & 0.0205 & 0.0019\\
    \hline
    \end{tabular}
    }
    \caption{Quantitative results of the dense pose estimation on ZJU\_MoCap dataset. Here, MSE is the mean squared error between the predicted and estimated Continuous Surface Embeddings for Dense Pose.}
    \label{tab:densepose}
\end{table}
\subsection{Performance on dense human pose estimation}
\begin{figure}[ht!]
    \centering
     \includegraphics[width=1\linewidth]{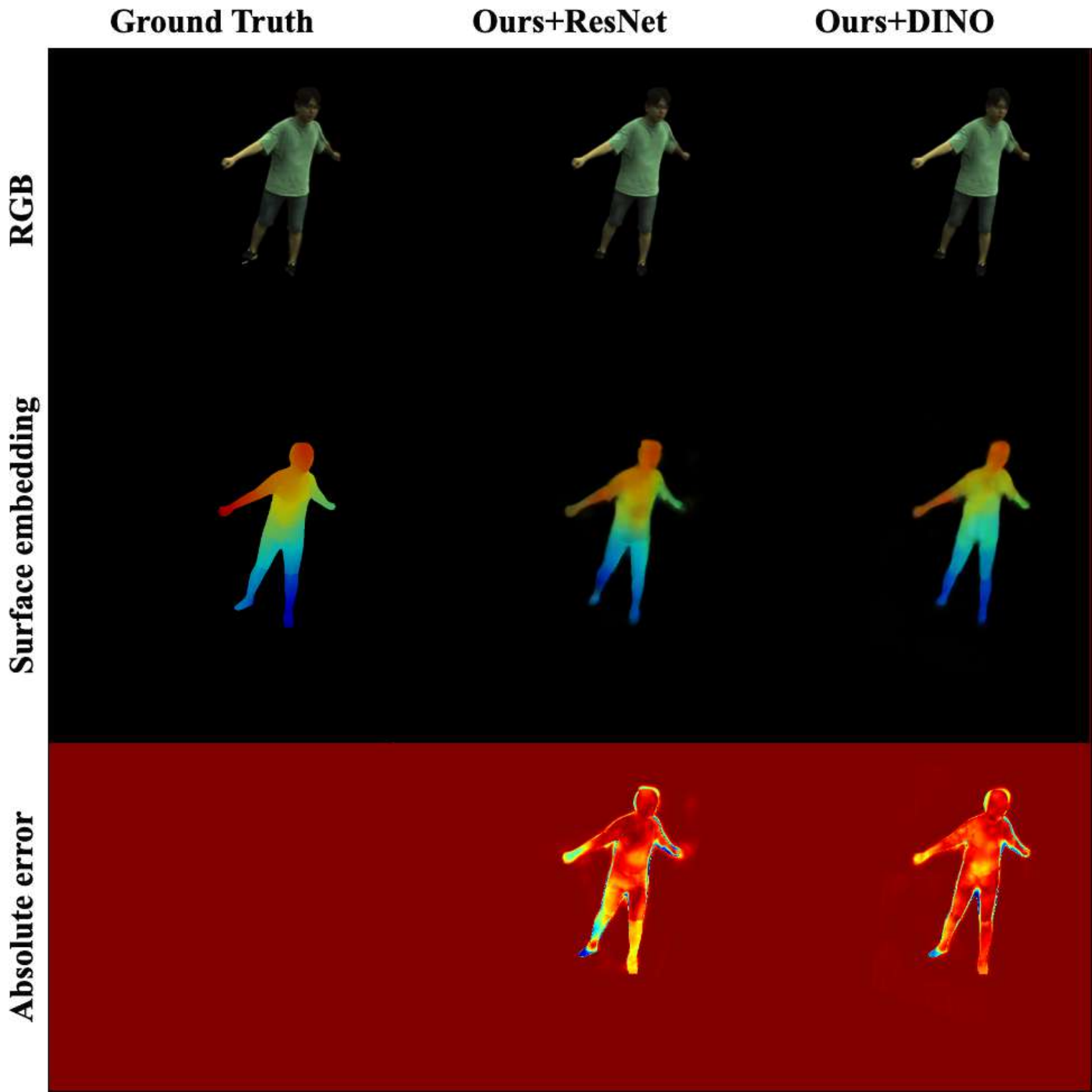}
    \caption{\small Qualitative result of dense pose estimation on ZJU\_MoCap dataset. The absolute error demonstrates the effectiveness of our model with DINO feature in learning dense pose.}
    \label{fig:dense}
\end{figure}

In order to showcase \shortname's ability to learn other generalizable human features, we conducted additional experiments to predict dense pose. During training, we use DensePose~\cite{guler2018densepose} to generate ground-truth Continuous Surface Embeddings of ZJU\_MoCap dataset. We used the same architecture without any modification to learn Continuous Surface Embeddings as human feature from 2D images, which can be used for dense pose estimation. We provide the quantitative results in Table~\ref{tab:densepose}.
The results show that our model can effectively estimate dense pose with different encoders, \eg, ResNet, and DINO, and we find that the
DINO encoder performs better compared to ResNet for dense pose estimation similar to joint estimation task. The qualitative results of the estimation of dense pose are presented in Fig.~\ref{fig:dense}. Both qualitative and quantitative findings demonstrate that \shortname is capable of learning other generalizable human features beyond just keypoint estimation. This experiment validates our assumption that \shortname can learn different human features using the same model architecture.
\subsection{Rendering speed} Inference time of various methods in novel view synthesis and keypoint estimation is illustrated in Table \ref{tab:rendering_speed}. We compared the proposed method with the baseline approach (ENeRF with an extra heatmap branch). While the utilization of the DINO encoder may result in longer inference times, it surpasses other methods by providing superior joint estimation. It may be feasible to attain faster inference time by employing a custom Visual Transformer-based encoder and optimization while maintaining the same level of performance. All experiments were performed on a single RTX 3090 GPU using the PyTorch implementation. We are confident that by optimizing and fine-tuning the code, the rendering time can be improved in the future.  
\begin{table}[ht!]
    \centering
    \small
    \begin{tabular}{c|c}
    \hline
         Method & FPS \\
    \hline
         ENeRF & 31.10 \\
         ENeRF+Heatmap & 27.81\\
         Ours+ResNet18 & 11.22\\
         Ours+ResNet34 & 10.49\\
         Ours+DINO & 4.08 \\
    \hline
    \end{tabular}
    \caption{Average rendering speed in FPS(Frame per second). ENeRF+Heatmap represent the baseline method.}
    \label{tab:rendering_speed}
\end{table}

\subsection{Ablation study}
In Table~\ref{tab:ablation}, we present the impact of different encoder architectures on the human joint estimation task. We have chosen two different encoder architectures inspired by previous state-of-the-art pose estimation algorithms, namely ResNet~\cite{he2016deep} and DINO~\cite{caron2021emerging}. Both methods produced comparable results in terms of visual quality, but DINO outperformed significantly in the joint estimation task. 
\begin{table}[tbh!]
    \centering
    \small
    \scalebox{0.87}{
    \begin{tabular}{c|ccccc}
    \hline
         Encoder & PSNR$\uparrow$ & SSIM$\uparrow$ & LPIPS$\downarrow$ & MSE$\downarrow$ &PCK$\uparrow$ \\
    \hline
         ResNet34 \textit{Pre}~\cite{he2016deep} & 31.53 & 0.965 & \textbf{0.049} & 0.0005 & 0.454\\
         ResNet34 \textit{Fine}~\cite{he2016deep} & 31.20 & 0.963  &  0.054 & 0.0004& 0.573\\
         DINO \textit{Pre}~\cite{caron2021emerging} & 31.28 & 0.964 & 0.051 & 0.0003 & 0.682\\
         DINO \textit{Fine}~\cite{caron2021emerging} & \textbf{31.61} & \textbf{0.966} & 0.050 & \textbf{0.0003} & \textbf{0.691}\\
    \hline
    \end{tabular}
    }
    \caption{Ablation study for keypoint estimation. We show a comparison between different types of encoder for keypoint estimation task. We evaluated both models on ZJU\_MoCap dataset. \textit{Pre} represents Pre-trained and \textit{Fine} denotes Finetune during the training.}
    \vspace{-10pt}
    \label{tab:ablation}
\end{table}

\section{Conclusion}
In this paper, we present \shortname an end-to-end framework to learn generalizable NeRF to estimate human biomechanic features from 2D images. Through extensive experiments, we have established that our approach can be successfully applied in a variety of settings. We addressed the shortcomings of underlying structure in previous NeRF based methods for humans. The proposed method utilizes an encoder to predict human features using NeRF. In this paper, we focus on estimating human keypoints, and we have also shown how it can be extended to other human features by estimating dense pose.
Although our method can estimate human features efficiently, it still has the following shortcomings: 1. It only works in scenes with a single human and it cannot handle multiple humans. 2. The proposed method is limited to humans and does not apply to other animals and articulated objects, which can be a future perspective to learn more general underlying structure. 
\section{Acknowledgement}
This project has received funding from the H2020 COFUND program BoostUrCareer under MSCA no.847581. 
{
    \small
    \bibliographystyle{ieeenat_fullname}
    \bibliography{main}
}

\pagebreak
\clearpage
\normalsize
\begin{center}
\textbf{\large Supplementary Material: \longname}
\end{center}

\setcounter{equation}{0}
\setcounter{figure}{0}
\setcounter{table}{0}
\setcounter{page}{1}
\setcounter{section}{0}
\makeatletter
\renewcommand{\theequation}{S\arabic{equation}}
\renewcommand{\thefigure}{S\arabic{figure}}
\renewcommand{\bibnumfmt}[1]{[#1]}
\renewcommand{\citenumfont}[1]{#1}

\section{Implementation Details}
In the following section, we present details regarding implementation to ensure reproducibility. It is important to note that we did not extensively optimize the architecture or the training procedure due to the significant computational time and resource needed. Thus, there is the possibility that different variations of the hyperparameter can result in a better model.
\subsection{Human Feature Encoder} 
This section presents a comprehensive explanation of our human feature encoder, as introduced in the main paper. Our approach integrates two encoder architectures: DINO and ResNet with a focus on the ResNet34 and ResNet18 variants. For an image with dimensions $H \times W$, the feature map obtained from the ResNet encoder has dimensions $512 \times H \times W$. Following the approach of PixelNeRF~\cite{yu2021pixelnerf}, we utilized a pre-trained ResNet model on ImageNet. We extracted a feature pyramid similar to PixelNeRF and concatenated them to generate a feature volume of size $512 \times H/2 \times W/2$. Finally, the feature volume was upsampled to generate a human feature with dimensions $512 \times H \times W$. 
\par Additionally, our framework incorporates a pre-trained DINO~\cite{caron2021emerging} ViT-Small model with a patch size of 8, which was obtained from the official GitHub repository. To generate the final feature, we extracted features from the 9th and 11th layers of the DINO model and concatenated them. The resulting feature has a shape of $384 \times H/8 \times W/8$. Subsequently, we up-sampled the features using bilinear interpolation to obtain a feature shape of $384 \times H \times W$.

\subsection{NeRF Architecture}We utilized an MLP named $g_{NeRF}$ to produce intermediate NeRF features from images. Subsequently, smaller MLPs were used to generate the outputs. In our experiments, we used 2 fully connected layers for $g_{NeRF}$, which takes $f_{img}$ and $f_{voxel}$ as inputs, following a similar approach as described in ENeRF~\cite{lin2022efficient}. To generate the density $\sigma$ from the intermediate NeRF feature, we used an MLP $g_{\sigma}$, which consists of a single linear layer followed by a softmax layer. To estimate the pixel color, we used an MLP $g_c$ with 2 linear layers and 2 ReLU activation functions. The heatmap generation was performed using the $g_h$ MLP, which takes $V_{NeRF}$ and $f_h$ as input. $g_h$ utilizes 2 linear layers with ReLU and Sigmoid activation functions.

\subsection{Experimental Setup}
We assessed the performance of our method using two datasets: ZJU\_MoCap and RenderPeople. For the ZJU\_MoCap dataset, we utilized 6 dynamic sequences, namely \emph{CoreView\_315, CoreView\_377, CoreView\_387, CoreView\_390, CoreView\_394,} and \emph{CoreView\_393} for training, and \textit{CoreView\_313 and CoreView\_386} for testing. We did not include the \emph{CoreView\_392} sequence in our evaluation, as it is missing frame data. We used the 2D joint locations provided by ZJU\_MoCap to generate heatmaps during training. For training and testing, we limited the frames to an initial 600 frames and divided the total number of cameras equally for training and testing purposes. During training and testing, we generate a 3D bounding box around the dynamic entity using the SMPL model provided in ZJU\_MoCap dataset, we project it to obtain a bounding mask and make the colors of pixels outside the mask as zero. For RenderPeople, we used the foreground mask of the dataset. Rays are sampled only inside the mask regions. We calculate PSNR, SSIM, and LPIPS within the masked region. For the RenderPeople dataset, we randomly chose 440 sequences for the training set and 60 sequences for the test set. As RenderPeople does not provide any keypoint information, we used OpenPose as a teacher network to learn the heatmap feature. We utilized 8 samples and 32 volume planes for the course network in all of our experiments, while for the fine network, we used 4 samples and 8 volume planes. We implemented our method and baseline with PyTorch. We report the evaluation metrics and the rendering speed using a single RTX 3090 GPU. We plan to incorporate more datasets for the purpose of benchmarking in future.

\section{Additional Experiments and Results}
In this section, we present additional results and experiments. 

\subsection{Coordinate Loss Function} To enhance the spatial perception of our NeRF representation, we introduce a coordinate loss, $l_{coord}$, aimed at minimizing the Mean Squared Error (MSE) between the input 3D coordinates and the 3D points regressed by the network. This is achieved by incorporating an additional branch in the output to approximate the input query point $x$. The MLP $g_{co}$, responsible for this task, processes intermediate NeRF features through a linear layer followed by a ReLU activation function. The composite loss function is formulated as follows:
\begin{equation}
l = l_{col} + \lambda_p l_{perc} + \lambda_h l_{heat} + \lambda_c l_{coord}
\end{equation}
where the weighting coefficients $\lambda_p$, $\lambda_h$, and $\lambda_c$ are set to 0.01, 0.5, and 0.01/0.05, respectively. Table~\ref{tab:coordinate} shows the quantitative results of our experiments with coordinate loss. 
\begin{table*}[h!]
    \centering
    \small
    \begin{tabular}{c|ccccc}
    \hline
         Encoder & PSNR & SSIM & LPIPS & MSE &PCK \\
    \hline
         ResNet34 & 31.20 & 0.963  &  0.054 & 0.0004& 0.573\\
         DINO & 31.61 & 0.966 & 0.050 & 0.0003 & 0.687\\
         ResNet34+co+0.01 & 31.57 & 0.964 & 0.057& 0.0010& 0.292\\
         ResNet34+co+0.05 & 31.33 & 0.958 &0.072 & 0.0008 & 0.427\\
    \hline
    \end{tabular}
    \caption{Quantitative results of coordinate loss experiments compare to other methods. }
    \label{tab:coordinate}
\end{table*}

\subsection{Additional results}
In this section, we further explore the qualitative and quantitative results obtained from the ZJU\_MoCap and RenderPeople datasets.\\
\subsubsection{ZJU\_MoCap dataset:} 
\begin{figure*}[ht!]
    \centering
     \includegraphics[width=1\linewidth]{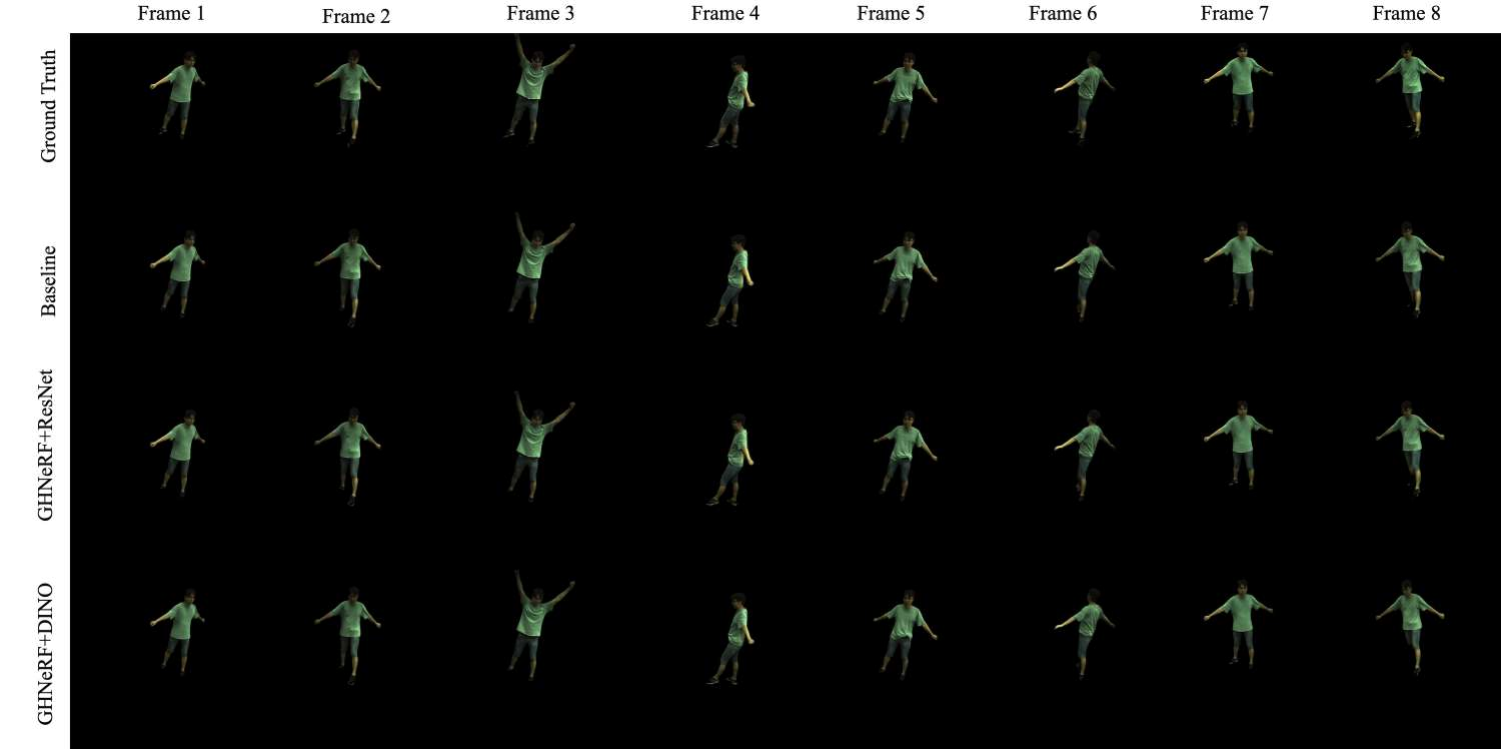}
    \caption{Qualitative results on \emph{CoreView\_313} sequence of ZJU\_MoCap dataset.}
    \label{fig:sr1}
\end{figure*}

\begin{figure*}[ht!]
    \centering
     \includegraphics[width=1\linewidth]{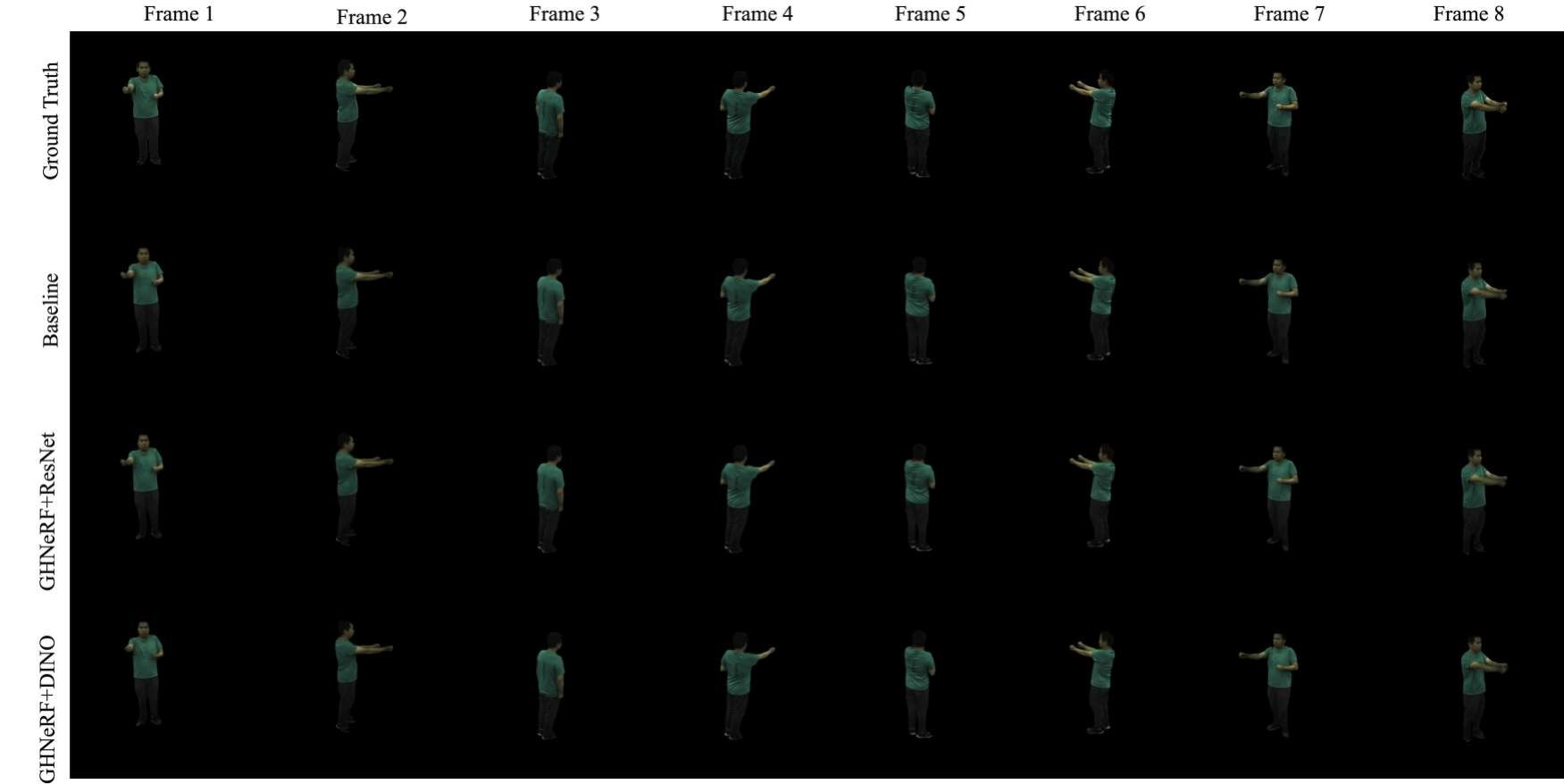}
    \caption{Qualitative results on \emph{CoreView\_386} sequence of ZJU\_MoCap dataset.}
    \label{fig:sr2}
\end{figure*}

\begin{figure*}[ht!]
    \centering
     \includegraphics[width=1\linewidth]{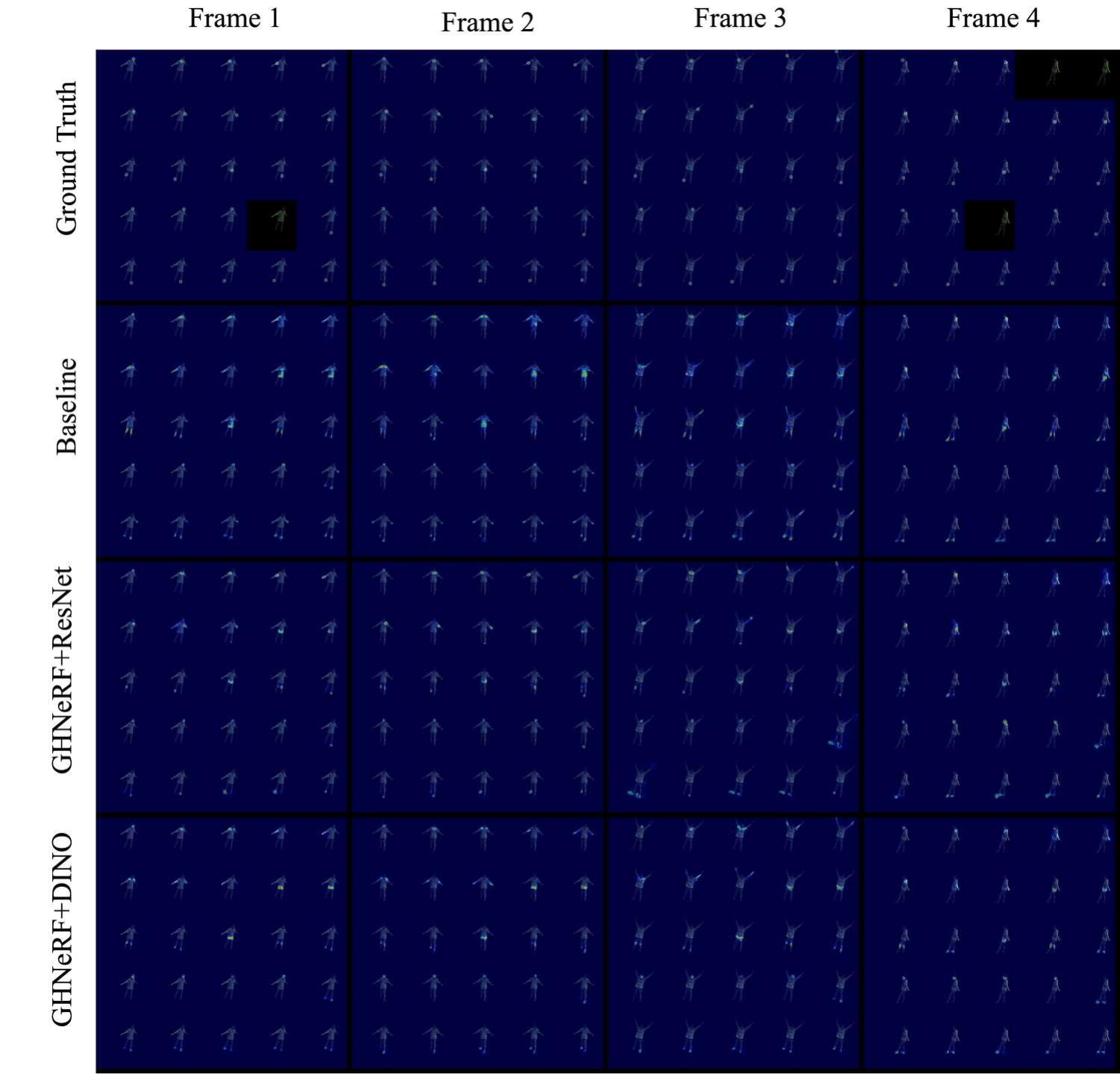}
    \caption{Qualitative results of heatmap prediction on \emph{CoreView\_313} sequence of ZJU\_MoCap dataset. We estimated 25 keypoints and visualized each channel separately in $5 \times 5$ grids.}
    \label{fig:h1}
\end{figure*}

\begin{figure*}[ht!]
    \centering
     \includegraphics[width=1\linewidth]{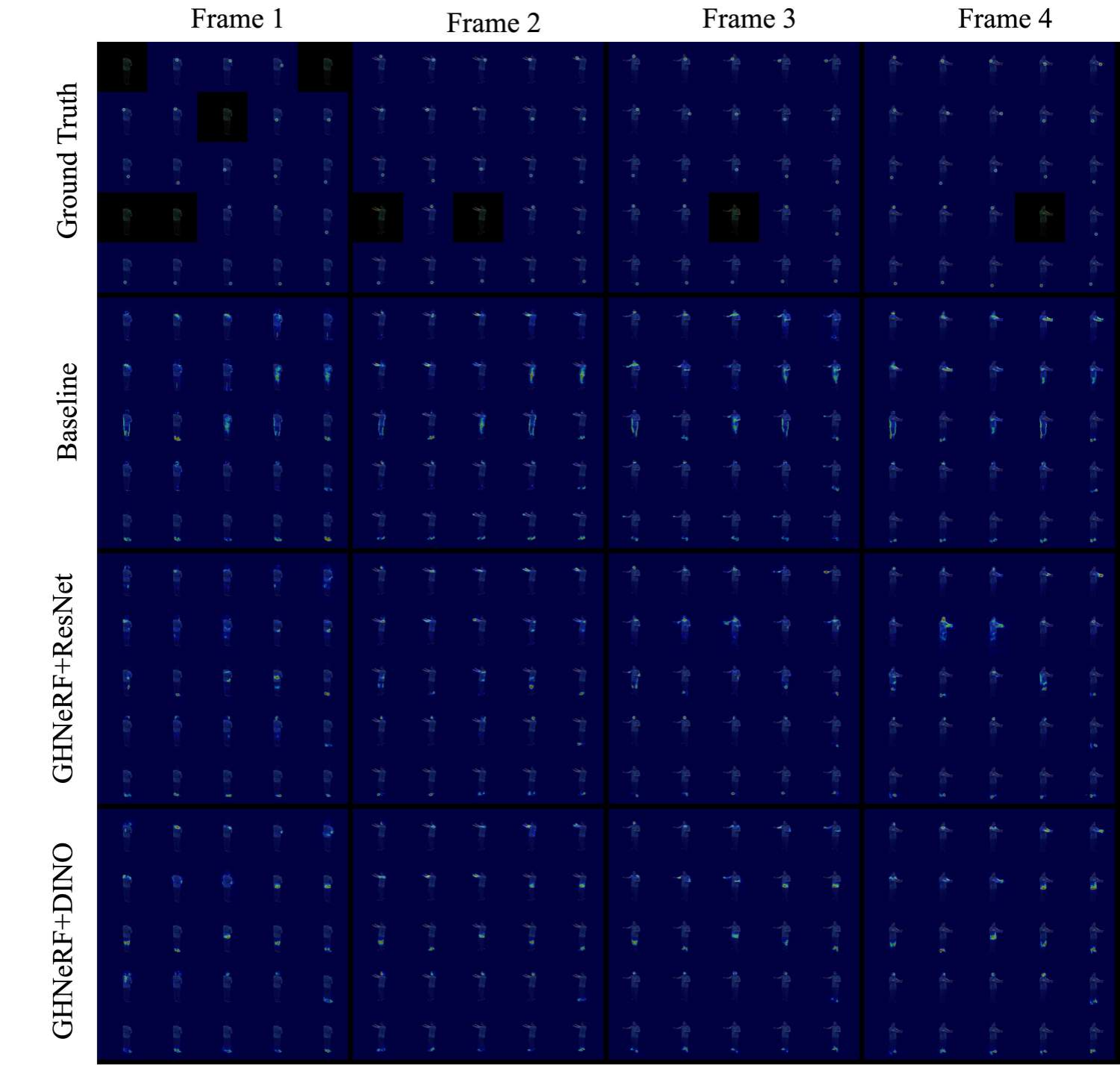}
    \caption{Qualitative results of heatmap prediction on \emph{CoreView\_386} sequence of ZJU\_MoCap dataset. We estimated 25 keypoints and visualized each channel separately in the $5 \times 5$ grids.}
    \label{fig:h2}
\end{figure*}

\begin{figure*}[ht!]
    \centering
     \includegraphics[width=1\linewidth]{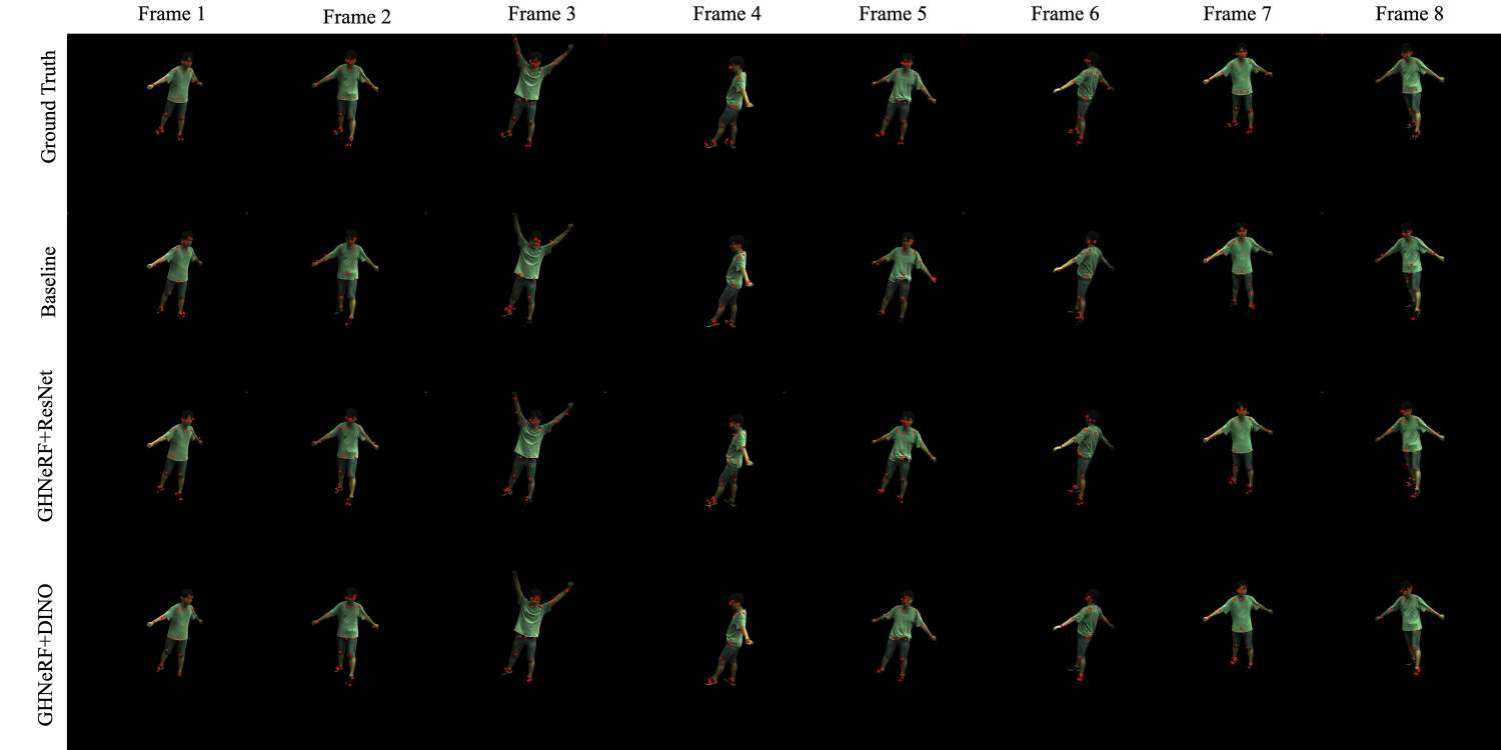}
    \caption{Qualitative results of keypoint estimation on \emph{CoreView\_313} sequence of ZJU\_MoCap dataset.}
    \label{fig:sk1}
\end{figure*}

\begin{figure*}[ht!]
    \centering
     \includegraphics[width=1\linewidth]{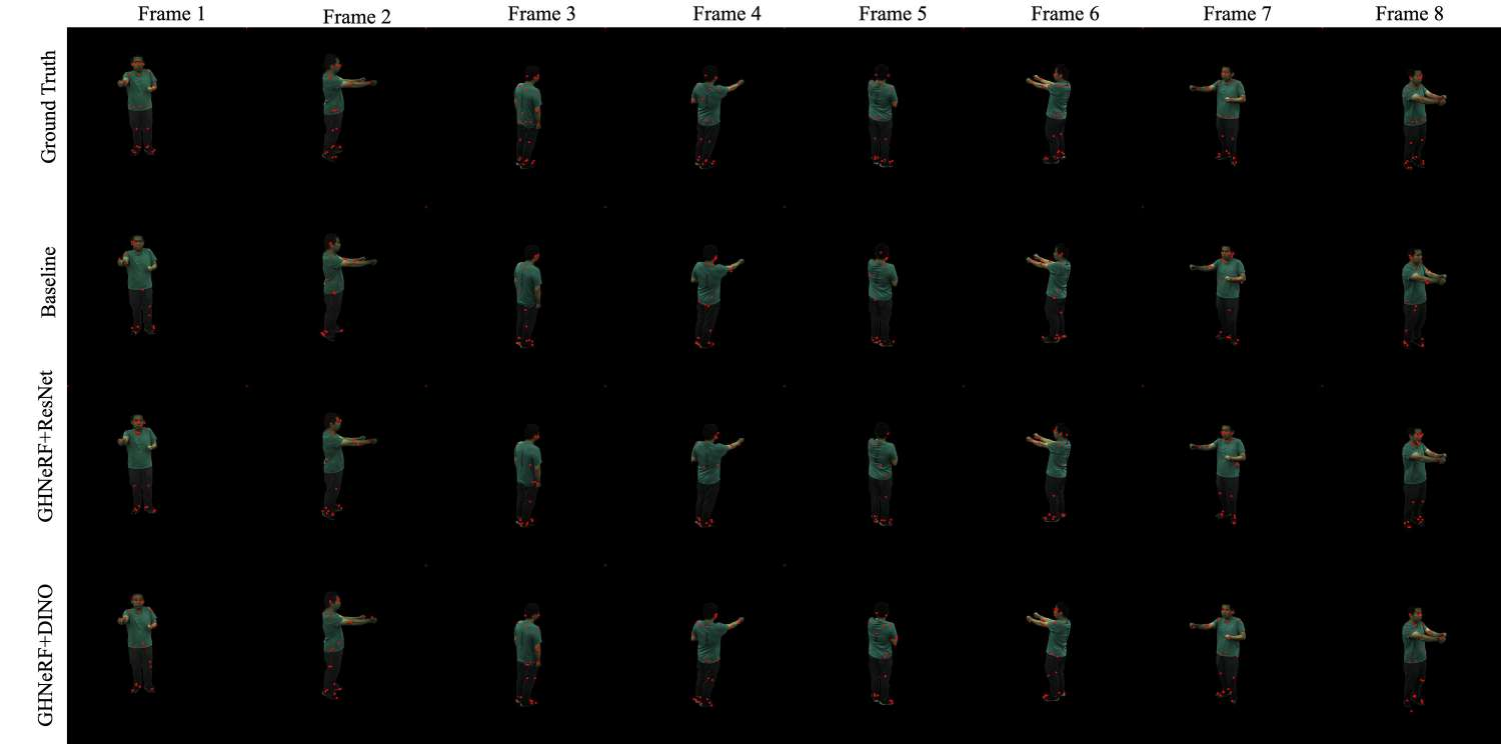}
    \caption{Qualitative results of keypoint estimation on \emph{CoreView\_386} sequence of ZJU\_MoCap dataset.}
    \label{fig:sk2}
\end{figure*}

Additional qualitative insights for the novel view synthesis on ZJU\_MoCap are illustrated in Figures~\ref{fig:sr1} and \ref{fig:sr2}. Our proposed method, \shortname, uses heatmaps for keypoint estimation. The estimated heatmaps generated by our method are shown and compared in Figures \ref{fig:h1} and \ref{fig:h2}. We have observed missing data in the ground-truth heatmaps. To ensure accurate evaluation metrics, we have excluded the keypoints associated with these missing data. We have compared our 2D keypoint estimate with the baseline in Figures \ref{fig:sk1} and \ref{fig:sk2}.\\
\subsubsection{RenderPeople dataset:} 
\begin{figure*}[ht!]
    \centering
     \includegraphics[width=1\linewidth]{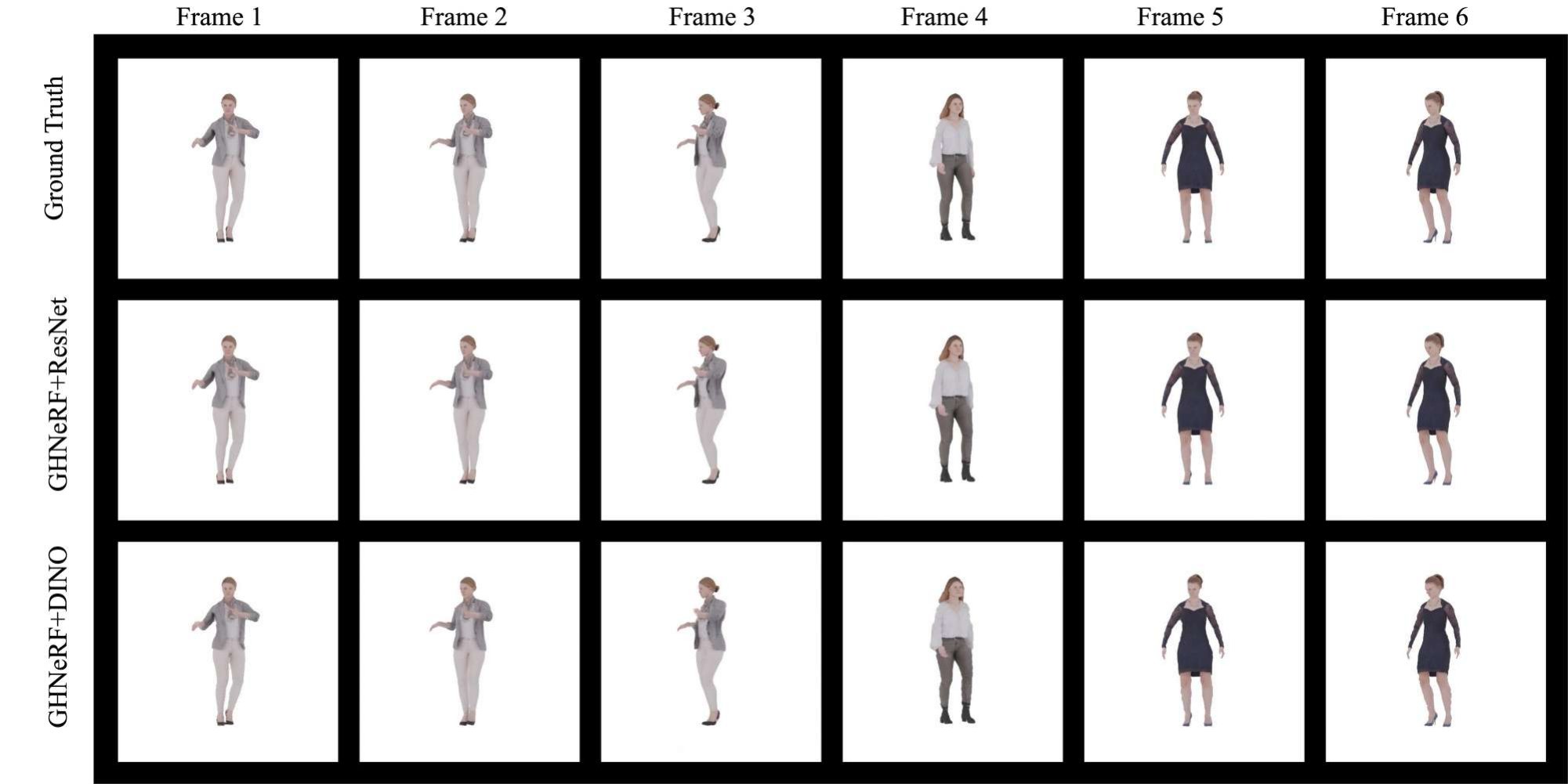}
    \caption{Qualitative results of novel view synthesis on RenderPeople dataset.}
    \label{fig:srr1}
\end{figure*}

\begin{figure*}[ht!]
    \centering
     \includegraphics[width=1\linewidth]{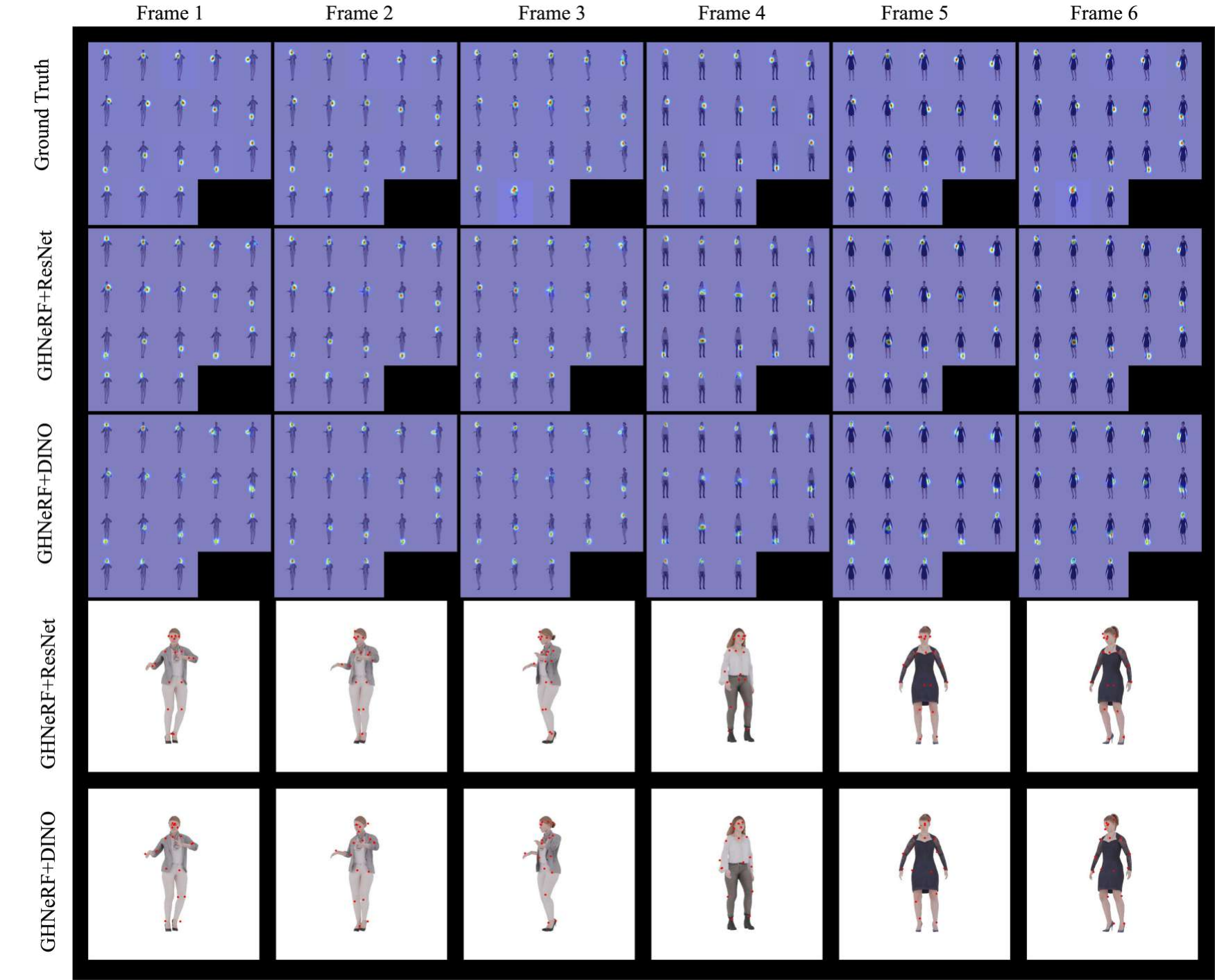}
    \caption{Qualitative results on RenderPeople dataset. The illustration shows predicted heatmaps along with estimated keypoints from heatmaps.}
    \label{fig:srk1}
\end{figure*}

In order to demonstrate the effectiveness of our approach on various human images, we evaluated its performance using the RenderPeople dataset, which is a simulated dataset. The RenderPeople dataset does not include any ground-truth keypoints, therefore, we train our model for the keypoint estimation task by distilling a state-of-the-art pose estimation algorithm. We provided qualitative results of the novel view synthesis on the RenderPeople dataset in Figure \ref{fig:srr1}. In Figure \ref{fig:srk1}, we present the performance of our model in heatmap estimation and keypoint prediction. We used an image resolution of $512 \times 512$ for all experiments conducted on the RenderPeople dataset.
\subsubsection{Dense Pose estimation:}
\begin{figure*}[ht!]
    \centering
     \includegraphics[width=0.9\linewidth]{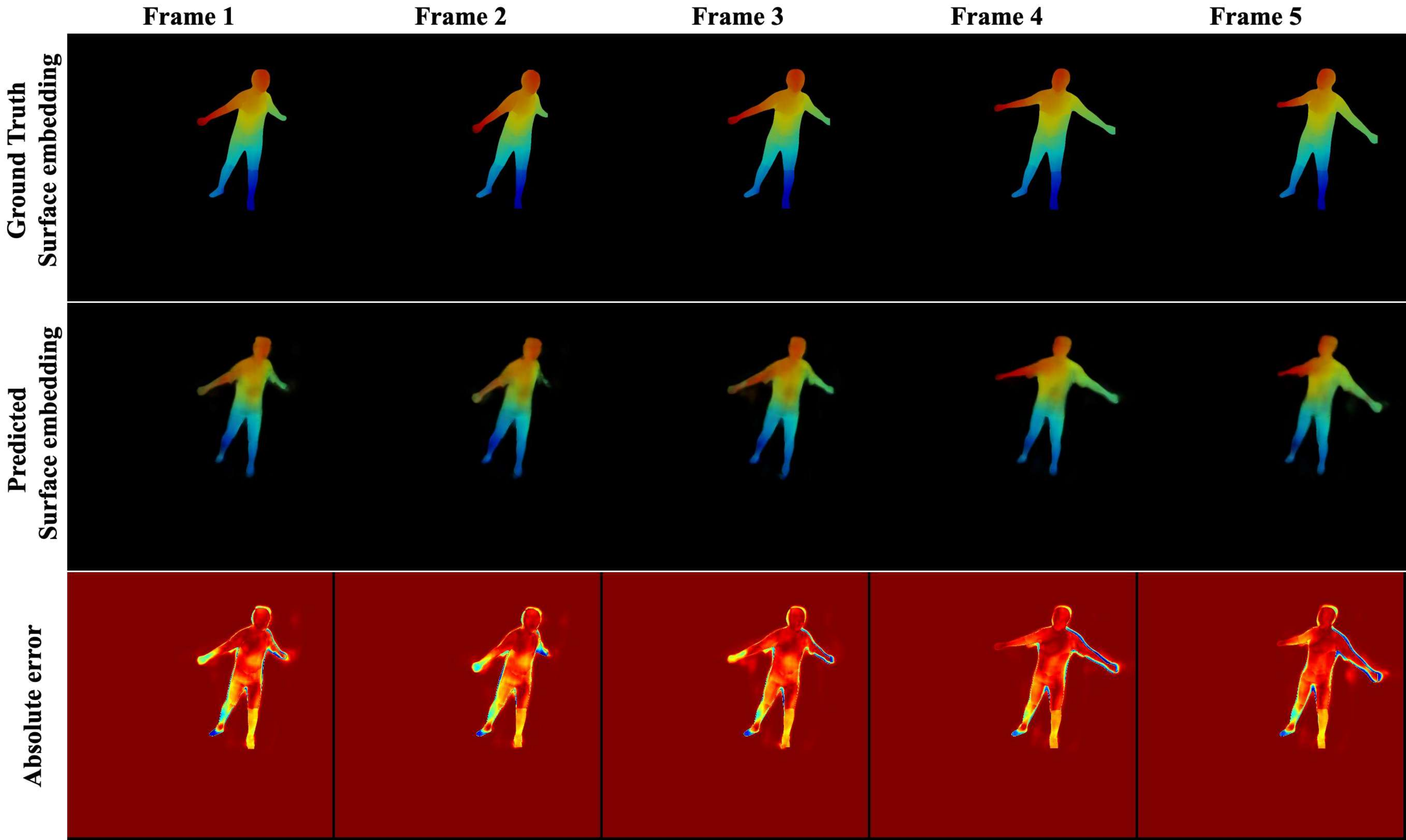}
    \caption{Qualitative results of dense pose estimation with ResNet encoder. We have compared ground truth and predicted Continuous Surface Embeddings.}
    \label{fig:dr}
\end{figure*}

\begin{figure*}[ht!]
    \centering
     \includegraphics[width=0.9\linewidth]{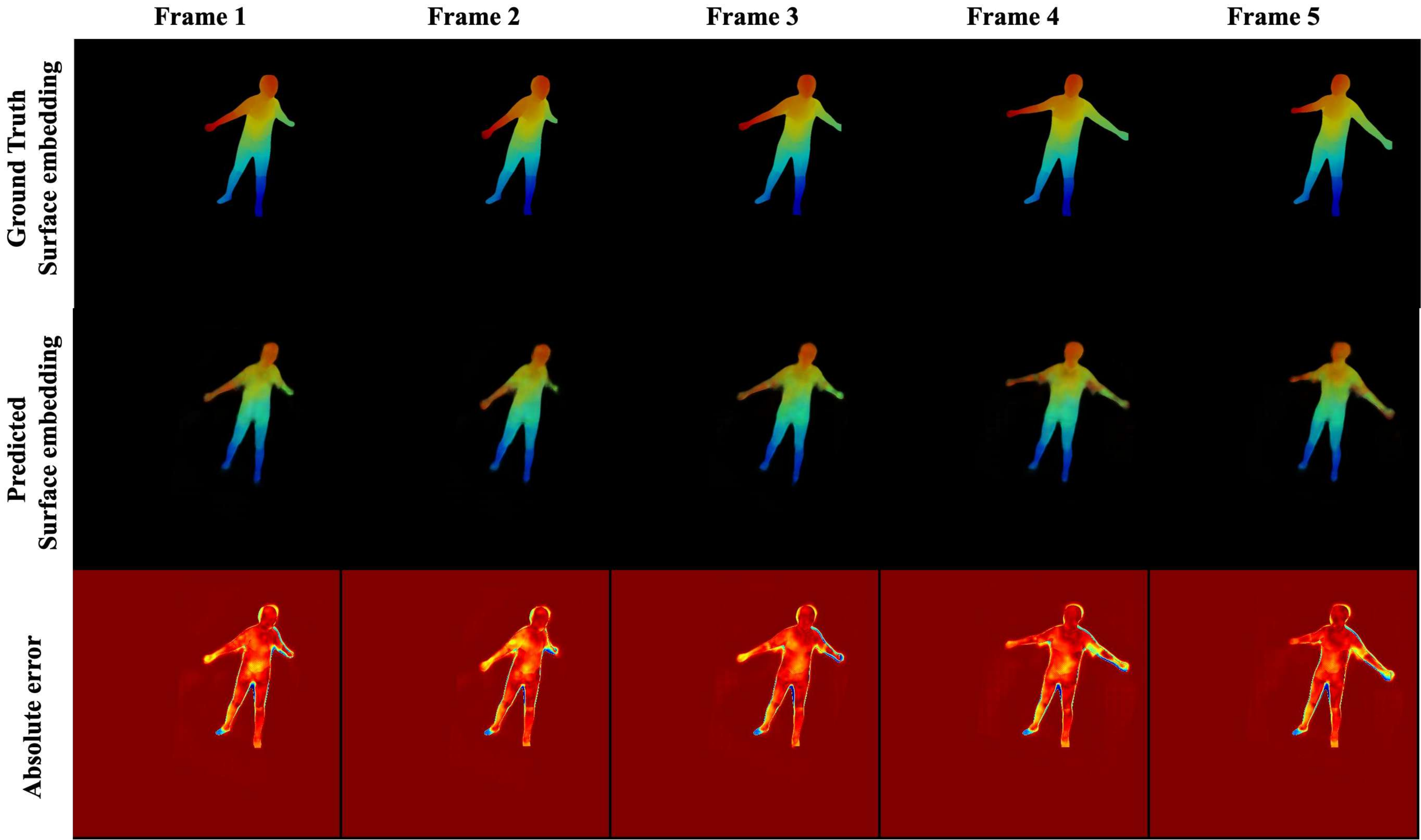}
    \caption{Qualitative results of dense pose estimation with DINO encoder. We have compared ground truth and predicted Continuous Surface Embeddings.}
    \label{fig:dd}
\end{figure*}

We conducted additional experiments to demonstrate that \shortname can be utilized to estimate various human features beyond just keypoints. Our model was trained on ZJU\_MoCap dataset to predict dense human pose as Continuous Surface Embedding. We trained our model by distilling the SoTA DensePose\cite{guler2018densepose} algorithm. We have presented the qualitative results of dense pose estimation with the ResNet and DINO encoder in Figure~\ref{fig:dr} and Figure~\ref{fig:dd}, respectively.


\end{document}